\documentclass[letterpaper]{article} 
\usepackage{aaai2025}  
\usepackage{times}  
\usepackage{helvet}  
\usepackage{courier}  
\usepackage[hyphens]{url}  
\usepackage{graphicx} 
\usepackage{natbib}  
\usepackage{caption} 
\usepackage{algorithm}
\usepackage{algorithmic}

\usepackage{multirow}
\usepackage{amssymb}
\usepackage[table]{xcolor}

\usepackage{newfloat}
\usepackage{listings}
\DeclareCaptionStyle{ruled}{labelfont=normalfont,labelsep=colon,strut=off} 
\floatstyle{ruled}
\newfloat{listing}{tb}{lst}{}
\floatname{listing}{Listing}
\title{Multimodal 3D Reasoning Segmentation with Complex Scenes}
\author{
    Xueying Jiang\textsuperscript{\rm 1},
    Lewei Lu\textsuperscript{\rm 2},
    Ling Shao\textsuperscript{\rm 3},
    Shijian Lu\textsuperscript{\rm 1}\thanks{Corresponding author.}
}
\affiliations{
    \textsuperscript{\rm 1}S-Lab, Nanyang Technological University, Singapore\\
    \textsuperscript{\rm 2}Sensetime Research, China\\
    \textsuperscript{\rm 3}UCAS-Terminus AI Lab, University of Chinese Academy of Sciences, China\\
}

\usepackage{bibentry}

\begin{document}

\maketitle

\begin{abstract}
The recent development in multimodal learning has greatly advanced the research in 3D scene understanding in various real-world tasks such as embodied AI. However, most existing studies are facing two common challenges: 1) they are short of reasoning ability for interaction and interpretation of human intentions and 2) they focus on scenarios with single-category objects and over-simplified textual descriptions and neglect multi-object scenarios with complicated spatial relations among objects. We address the above challenges by proposing a 3D reasoning segmentation task for reasoning segmentation with multiple objects in scenes. The task allows producing 3D segmentation masks and detailed textual explanations as enriched by 3D spatial relations among objects. To this end, we create ReasonSeg3D, a large-scale and high-quality benchmark that integrates 3D segmentation masks and 3D spatial relations with generated question-answer pairs. In addition, we design MORE3D, a novel 3D reasoning network that works with queries of multiple objects and is tailored for 3D scene understanding. MORE3D learns detailed explanations on 3D relations and employs them to capture spatial information of objects and reason textual outputs. Extensive experiments show that MORE3D excels in reasoning and segmenting complex multi-object 3D scenes. In addition, the created ReasonSeg3D offers a valuable platform for future exploration of 3D reasoning segmentation. The data and code will be released.
\end{abstract}

\section{Introduction}

With the recent advancements in deep learning, 3D scene understanding has become one key component in various real-world tasks such as embodied AI and autonomous driving. The capabilities of understanding spatial relations among 3D objects in scenes and grasping the intention of human users have become essential for machines to interpret 3D scenes, interact with 3D objects in scenes, and achieve various complex and real-world missions while navigating within 3D environments.

On the other hand, most existing 3D scene understanding work does not possess reasoning and interpretation abilities for interacting with user textual inputs. For example, several studies introduce foundation models such as large language models (LLMs) to empower 3D scene understanding on captioning~\cite{chen2023unit3d,chen2023end,hong20233d}, question answering~\cite{azuma2022scanqa, ma2023sqa3d, parelli2023clip, guo2023point, hong20233d}, visual grounding~\cite{huang2022multi, zhu20233d, guo2023viewrefer, yang2024llm, chen2024ll3da, hong20233d, kang2024intent3d}, and referring~\cite{qian2024x, wu20243dgres, huang2021text, yuan2021instancerefer}, but they are short of reasoning capabilities for deducing human intentions. Several recent studies attempt to introduce the reasoning ability of LLMs into 3D scene understanding tasks, but most of them focus on segmenting single or single-category objects and cannot handle complex scenes with multiple objects of different categories~\cite{chen2024ll3da, he2024segpoint,huang2025reason3d}. As a result, they cannot understand 3D spatial relations among objects and produce fine-grained textual explanations, hampering application in real-world scenarios with multiple objects of different categories.

\begin{figure*}[ht]
    \centering
    \includegraphics[width=1.0\linewidth]{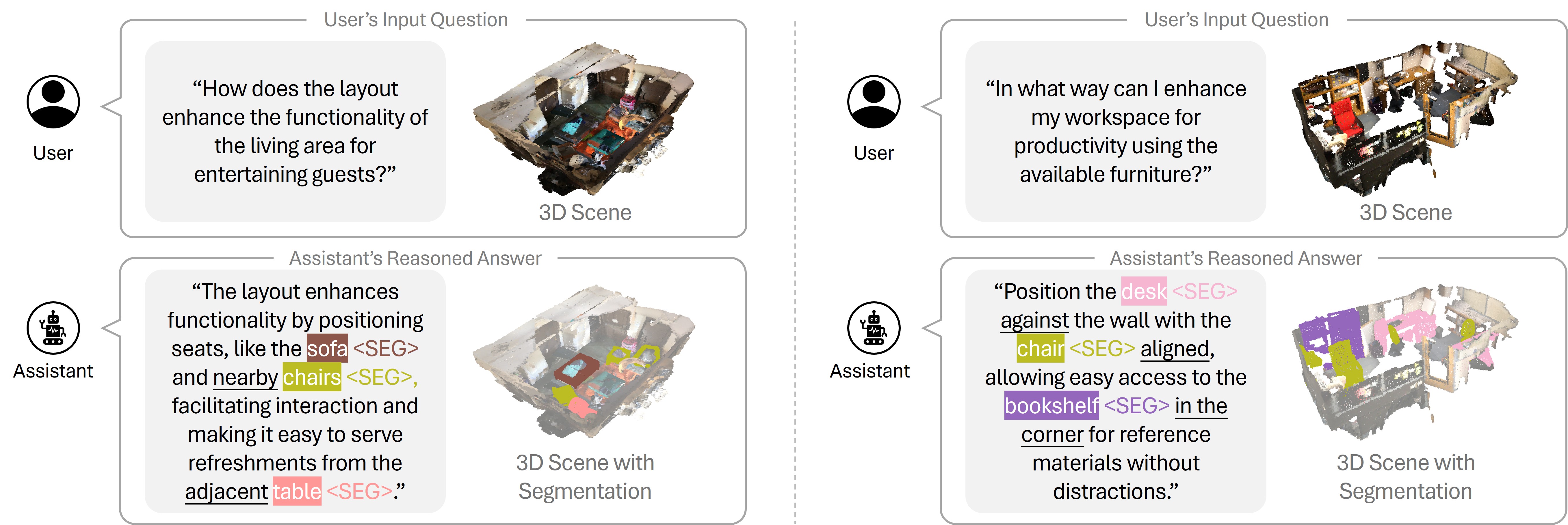}
    \caption{
    The proposed MORE3D enables multi-object 3D reasoning segmentation. It can comprehend the intention behind user questions or instructions, handle complex 3D scenes with multiple objects of different categories, and produce fine-grained explanations with 3D spatial relations among objects, demonstrating strong reasoning and 3D segmentation capabilities.
    }
    \label{fig:motivation}
\end{figure*}

We propose a multi-object 3D reasoning segmentation task that can produce 3D segmentation masks and textual explanations with rich 3D spatial relations among objects in scenes, given 3D scenes and user questions as inputs. The task consists of two key components. The first is ReasonSeg3D, a large-scale and high-quality benchmark that can evaluate 3D reasoning segmentation with multiple 3D objects and rich spatial relations among them. Different from existing 3D reasoning segmentation benchmarks~\cite{he2024segpoint,huang2025reason3d}, ReasonSeg3D expands the scope into multi-object space which is well aligned with real-world tasks that often come with multiple objects. In addition, ReasonSeg3D integrates 3D spatial information into question-answer pairs, where the 3D spatial relations in textual answers benefit 3D reasoning segmentation clearly. The second is  \textit{MORE3D}, a simple yet effective technique that enables multi-object reasoning segmentation with textual explanations to users’ questions. 
MORE3D extracts object-specific point cloud embedding for precise prediction of multi-category objects in complex 3D scenes. It incorporates textual explanations with detailed 3D spatial relations of multiple objects in answers, guiding the model toward a comprehensive understanding of 3D scenes. The generated textual answers support accurate segmentation and comprehensive reasoning for complex 3D scenes.
As illustrated in Figure~\ref{fig:motivation}, MORE3D demonstrates strong reasoning capability for comprehending the intention behind user's input questions and complex 3D scenes, producing accurate 3D segmentation and explanatory answers with respect to multiple 3D objects in scenes.

The major contributions of this work can be summarized in three aspects. First, we introduce a new multi-object 3D reasoning segmentation task, together with a large-scale and high-quality benchmark that incorporates 3D spatial relations for effective evaluations of multi-object 3D reasoning segmentation. Second, we design a reasoning segmentation technique that can handle multi-object 3D reasoning segmentation and produce fine-grained textual explanations with 3D spatial relations among objects in scenes. Third, extensive experiments demonstrate the superiority of our proposed multi-object reasoning segmentation technique as well as the validity of our created benchmark on multi-object 3D reasoning segmentation. 

\section{Related Work}

\subsection{Language-Instructed 3D Tasks}

Integrating point clouds with natural language processing has widespread applications, drawing increasing interest in language-instructed 3D scene understanding. Existing language-instructed 3D scene understanding methods can be broadly grouped into two categories. The first category focuses on 3D segmentation~\cite{nguyen2023open3dis, yan2024maskclustering, boudjoghra2024open, ding2023pla, yang2024regionplc, huang2023openins3d, liu2023weakly, qian2024x, wu20243dgres, huang2021text, yuan2021instancerefer, peng2023openscene, takmaz2023openmask3d, wu20243d, wu2024rg}, such as OpenScene~\cite{peng2023openscene},  Openmask3D~\cite{takmaz2023openmask3d}, 3D-STMN~\cite{wu20243d}, RG-SAN~\cite{wu2024rg}, which produce segmentation masks but lack reasoning abilities and textual or conversational output. This limits their ability to comprehend the user's intention and interact with the user in real-world applications. The second category focuses on tasks such as 3D captioning~\cite{chen2023unit3d, chen2023end, hong20233d, chen2024grounded}, 3D question answering~\cite{azuma2022scanqa, ma2023sqa3d, parelli2023clip, jiang2025exploring, guo2023point, hong20233d, chen2024grounded}, and visual grounding~\cite{huang2022multi, zhu20233d, guo2023viewrefer, yang2024llm, chen2024ll3da, hong20233d, kang2024intent3d, zhang2023multi3drefer, zhang2024task, zhu2024scanreason, chen2024grounded}. They can generate textual outputs like phrases or conversational outputs, but leave the fine-grained segmentation task untouched and have no reasoning ability either. Different from existing methods, the proposed MORE3D predicts textual answers with explanation and accurate segmentation of multiple 3D objects within complex scenes, demonstrating strong reasoning capability in comprehending the intention behind the user's input questions.

\begin{table*}[t]

\centering
\setlength{\tabcolsep}{5pt}
\begin{tabular}{l|c|cc|cccc}
\hline
\multirow{2}{*}{Dataset} & \multirow{2}{*}{Venue} & \multicolumn{2}{c|}{Scale} & \multicolumn{4}{c}{Data for Each QA Pair} \\ \cline{3-8} 
                       &                            &     \# Scene & \# QA Pair  & S Object  & M Objects        & M Categories        & Explanation        \\ \hline\hline
Instruct3D~\cite{he2024segpoint} & ECCV 2024 & 280 & 2,565 &  \checkmark &\checkmark & \texttimes & \texttimes    \\ 
Reason3D~\cite{huang2025reason3d} & 3DV 2025 & - & 2,484 & \checkmark  &\texttimes & \texttimes & \texttimes   \\ 
ReasonSeg3D (Ours) & -  & \textbf{1,513} & \textbf{20,113} & \checkmark  & \checkmark & \checkmark  &  \checkmark     \\  \hline
\end{tabular}
\caption{
    Comparison on 3D scene-level reasoning segmentation datasets. ReasonSeg3D stands out with large-scale, high-quality data, supporting segmenting multiple 3D objects across multiple categories and offering explanations with 3D spatial relations. S denotes single, and M denotes multiple.
    }
\label{tab:comparison_dataset}
\end{table*}

\subsection{Reasoning Segmentation}

Reasoning Segmentation is first introduced by LISA~\cite{lai2024lisa} to generate segmentation masks from complex, implicit textual queries. Specifically, LISA integrates LLaVA~\cite{liu2023visual} with SAM, enhancing segmentation through the vision-language model's reasoning capabilities. Following LISA, PixelLM~\cite{ren2024pixellm} improves pixel-level reasoning segmentation using multimodal models with a lightweight decoder and segmentation codebook, LLM-Seg~\cite{wang2024llm} bridges the Segmentation Anything Model and LLMs by selecting mask proposals, and LLaVASeg~\cite{yang2024empowering} incorporates query-focused segmentation into large language models where chain-of-thought prompting is adopted to preserve dialogue functions. In addition, VISA~\cite{yan2024visa} extends reasoning segmentation to video by combining multimodal language models with a mask decoder, facilitating complex video segmentation from implicit text queries and world knowledge. In the 3D domain, PARIS3D~\cite{kareem2024paris3d} and Reasoning3D~\cite{chen2024reasoning3d} focus on part segmentation with explanations for individual 3D objects, leaving 3D segmentation of complex scenes untouched. Recent studies like SegPoint~\cite{he2024segpoint} and Reason3D~\cite{huang2025reason3d} integrate LLMs' reasoning ability into 3D segmentation, but they are limited to segmentation for single-category objects and have no textual explanations. In contrast, our approach targets multi-object 3D segmentation and provides textual explanations in the output answer, enhancing the model's understanding of 3D spatial relations and offering a more practical solution for real-world complex scenes.

\subsection{Large Multimodal Models}

Inspired by the remarkable learning ability of Large Language Models (LLMs), recent research has expanded into the visual domain and developed a series of Large Multimodal Models (LMMs)~\cite{alayrac2022flamingo, zhang2023video}. 
The prevalent approach focuses on aligning visual representations with the linguistic embeddings of LLMs. For example, BLIP-2~\cite{li2023blip} and mPLUG-OWL~\cite{ye2023mplug} encode image features with a visual encoder, integrating them into the LLMs with text embeddings. LLaVA~\cite{liu2023visual} and MiniGPT4~\cite{zhu2024minigpt} align image-text features followed by instruction tuning, and they also explore image retrieval for LLMs.
Recent studies delve into the integration of multimodal LLMs with vision tasks. For example, VisionLLM~\cite{wang2023visionllm} provides an interface for vision-centric tasks via instruction tuning though it does not exploit LLMs for complex reasoning. VisionLLM-v2~\cite{wu2024visionllm} integrates visual perception, understanding, and generation within a unified framework by using a ``super link" to connect the multimodal large model with task-specific decoders. DetGPT~\cite{pi2023detgpt} introduces multimodal LLMs into open-vocabulary detectors for instruction-based detection tasks. GPT4RoI~\cite{zhang2023gpt4roi} introduces spatial boxes as inputs, training on region-text pairings. LISA~\cite{lai2024lisa} enhances segmentation in multimodal LLMs by introducing a \texttt{<SEG>} token. These existing studies primarily target downstream tasks in the 2D domain.
In contrast, our approach extends into the 3D domain to enable multi-object segmentation of 3D point clouds and provides textual explanations to support reasoning in the segmentation process.

\section{ReasonSeg3D Dataset}

Most existing reasoning segmentation datasets are not suitable for studying the proposed 3D multi-object reasoning segmentation task. 
Specifically, existing reasoning segmentation datasets have two critical limitations. First, the existing 2D reasoning segmentation datasets~\cite{lai2024lisa, ren2024pixellm, wang2024llm, yan2024visa} lack 3D data, making them unsuitable for 3D tasks. Second, the existing 3D reasoning datasets~\cite{he2024segpoint, huang2025reason3d} focus on single-category objects as illustrated in Table~\ref{tab:comparison_dataset}, and they contain only a few hundred scenes which are far fewer than standard 3D segmentation datasets~\cite{dai2017scannet, rozenberszki2022language}. Additionally, these 3D reasoning datasets lack textual explanations with spatial information, hindering them from training MLLMs for better 3D spatial relation understanding. We bridge this gap by proposing a data generation pipeline as well as a new 3D reasoning segmentation dataset ReasonSeg3D, with more details to be elaborated in the following subsections.

\subsection{Dataset Definition}

In the proposed ReasonSeg3D, each point cloud $P$ is paired with multiple $\{x_{que}, y_{ans}, M\}$ triplets, where $x_{que}$ is a question targeting one or more objects in the image, $y_{ans}$ is the textual answer containing the explanation for the reasoning segmentation, and $M$ represents the 3D segmentation masks corresponding to $y_{ans}$. The question $x_{que}$ is designed to require world knowledge and reasoning ability to accurately identify and segment multiple objects. For example, instead of directly asking ``\texttt{Where are the sofa and table?}", ReasonSeg3D formulates the question by ``\texttt{Where would be the most suitable place for reading a book in this layout?}". An answer $y_{ans}$ in ReasonSeg3D is constructed to involve multiple objects, and it also includes explanations with 3D spatial relations. For example, the $y_{ans}$ to the above question is formulated by ``\texttt{The corner with the sofa and a small table next to it can serve as a perfect reading nook, providing comfort and a quiet atmosphere.}".

\begin{figure*}[t]
	\centering
	\includegraphics[width=1.0\linewidth]{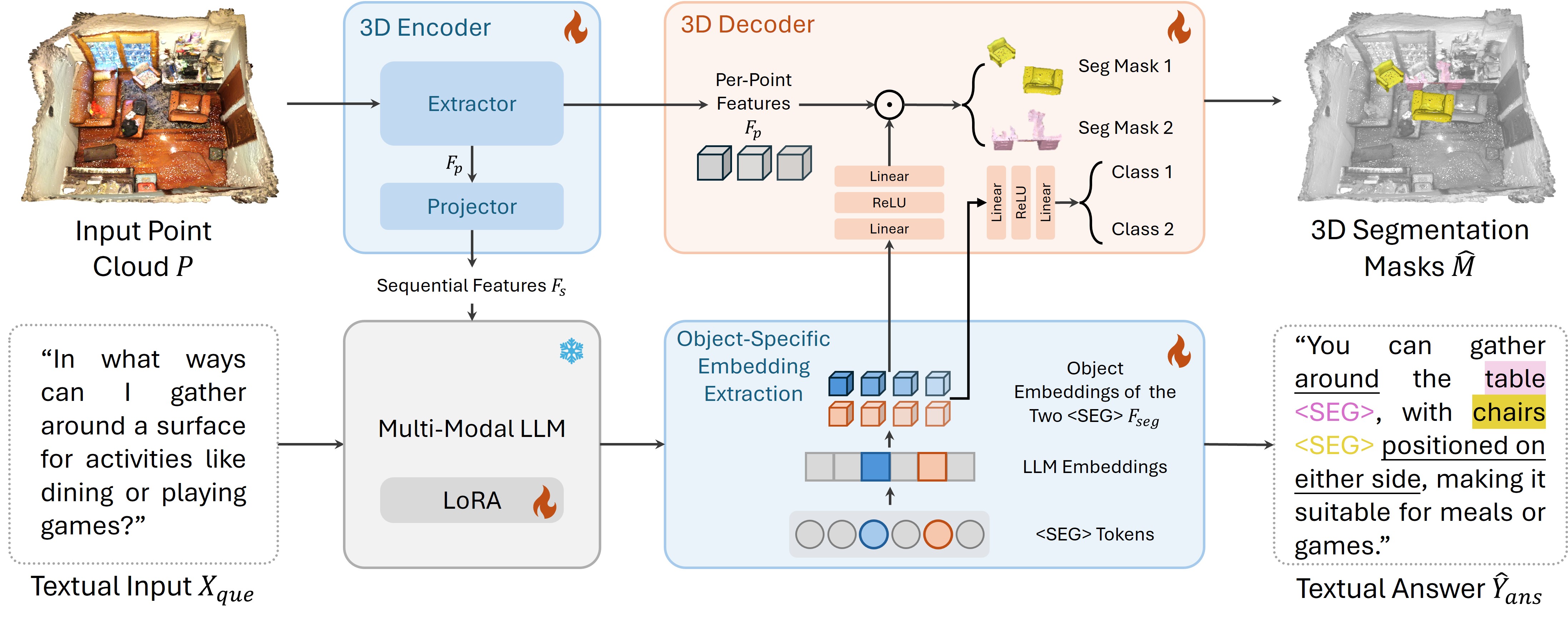}

	\caption{
    Overview of our proposed MORE3D: Given an input point cloud, the 3D Encoder first extracts per-point features $F_p$ and projects them into sequential features $F_s$. The sequential features $F_s$, together with the textual input $X_{que}$, are then fed into a multimodal LLM to perform reasoning, obtaining LLM embeddings and then producing textual answers $\hat{Y}_{ans}$ with both detailed explanations and descriptions of 3D spatial relations among multiple objects. Finally, after object-specific embedding extraction, object embeddings for multiple \texttt{<SEG>} tokens $F_{seg}$ and the per-point features $F_p$ are passed to the 3D Decoder to produce 3D segmentation masks. 
    Snowflake icon denotes frozen module, and flame icons denote trainable ones.
    }
	\label{fig:overall_architecture}
\end{figure*}

\subsection{Dataset Generation Pipeline}

Several reasoning segmentation datasets~\cite{lai2024lisa, wang2024llm} employ LLaVA~\cite{liu2023visual} for image captioning and text-only GPT-4 to produce question-answer pairs according to generated captions. However, GPT-4 cannot understand the image content well and its generated question-answer pairs often lack crucial 3D spatial relations that are essential for 3D reasoning segmentation in various real-world tasks.

We design a novel data generation pipeline that introduces GPT-4o which possesses superior visual content understanding capabilities. The pipeline allows the generation of more practical questions and answers by incorporating spatial relations among objects in scenes. Specifically, we employ scene image and ground-truth segmentation to enhance GPT-4o's 3D spatial understanding capabilities. The prompt template is structured in two parts: the first part contains basic requirements for the GPT-4o and the second specifies detailed requirements for generating questions and answers with a focus on describing 3D spatial relations among objects. With this template, GPT-4o autonomously selects objects to form question-answer pairs that reflect the scene’s content and spatial layout. 
On top of that, we conduct human verification to eliminate errors, such as incorrect object references and spatial‑relation errors, from the generated question–answer pairs. 
Since GPT-4o demonstrates excellent generation quality, only about 3\% of the generated question–answer pairs require manual correction.
Samples that maximize diversity in both target objects and spatial‑relation descriptions are then selected, thereby avoiding bias and ensuring high‑quality, varied question–answer pairs.

\subsection{Dataset Statistics}

ReasonSeg3D comprises 1,513 scenes and 20,113 data samples in total, with point clouds sourced from ScanNetv2~\cite{dai2017scannet}. Following \cite{dai2017scannet}, we divide ReasonSeg3D into training and validation sets, comprising 1,201 and 312 scenes, respectively. 
In ReasonSeg3D, each scene has 13.3 questions on average, with 20 different object categories for segmentation. Our dataset focuses on indoor scenes, which are critical for a wide range of real-world embodied applications, such as household robotics and augmented reality, where accurate reasoning and perception in complex, cluttered environments are essential.

\section{Method}

\subsection{Task Definition}

Multi-object 3D reasoning segmentation takes a single point cloud $P$ and input user questions $X_{que}$ as input, aiming to reason the implicit intention behind $X_{que}$ and produce textual answers $\hat{Y}_{ans}$ including explanations and 3D segmentation masks $\hat{M}$ of multiple objects in $P$.

\begin{figure*}[t]
    \centering
    \includegraphics[width=1.0\linewidth]{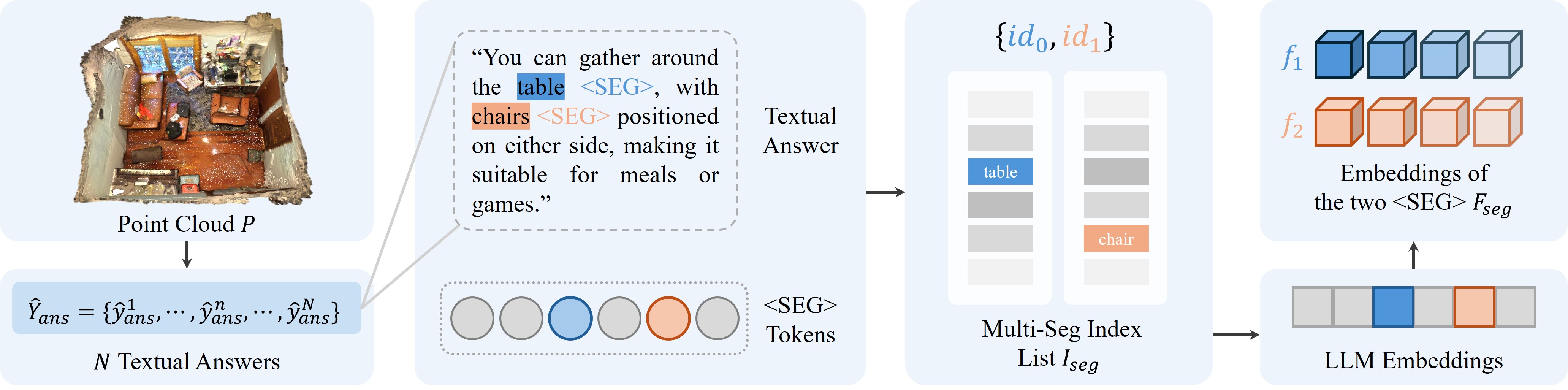}

    \caption{
    Extraction of object-specific point cloud embeddings. Each predicted textual answer contains multiple \texttt{<SEG>} tokens, with their positions recorded in a multi-segment index list $I_{seg}$. The LLM embedding corresponding to each \texttt{<SEG>} token is extracted based on the multi-segment index list for obtaining object-specific point cloud embeddings.
    }
    \label{fig:object_specific_feature_extraction}
\end{figure*}

\subsection{Overall Framework}

Figure~\ref{fig:overall_architecture} shows the framework of the proposed MORE3D which enables \textbf{M}ulti-\textbf{O}bject 3D \textbf{R}easoning segmentation and provides textual \textbf{E}xplanations based on the user's input question. Given an input point cloud $P$, the 3D Encoder first groups and extracts per-point features $F_p$ and then projects $F_p$ into sequential features $F_s$ that can be processed by LLMs. The sequential features $F_s$ and input text queries are processed by the LLM, obtaining LLM embeddings and then generating textual answers $\hat{Y}_{ans}$ that include textual explanations $\hat{Y}_{text}$ and multi-object \texttt{<SEG>} tokens $\hat{Y}_{\texttt{<SEG>}}$. The \texttt{<SEG>} tokens $\hat{Y}_{\texttt{<SEG>}}$ indicate the request for 3D segmentation masks, and the corresponding object embeddings $F_{seg}$ are extracted from the LLM embeddings. These object embeddings $F_{seg}$ are further combined with per-point features $F_p$ within 3D Decoder by dot product, ensuring that the object-specific information interacts with the overall point cloud features to generate the final 3D segmentation masks $\hat{M}$ and classification predictions.

\paragraph{Remark 1.} 

Different from prior studies on 2D reasoning segmentation~\cite{lai2024lisa, ren2024pixellm, wang2024llm} relying solely on 2D representations, which cannot effectively capture 3D geometric information or spatial relations in complex 3D scenes, MORE3D is tailored for 3D tasks and features a 3D-specific encoder-decoder for capturing complex 3D information. In addition, MORE3D extracts object-specific point cloud embeddings to handle multiple objects in 3D scenes, resolving unique challenges in 3D reasoning segmentation effectively. Further, MORE3D generates textual explanations that incorporate 3D spatial relations among objects, enabling explainable and spatially grounded reasoning of 3D scenes.

\subsection{Multi-Object 3D Reasoning Segmentation}
\label{sec:multi_object_3d_reasoning}

We design an object-specific embedding extraction approach to achieve multi-object 3D reasoning segmentation. This approach enables precise segmentation and reasoning for multiple objects of different categories in complex point clouds, integrating the LLM with a 3D point cloud decoder to generate segmentation masks and textual explanations.

The LLM takes point cloud features $F_s$ and text queries $X_{que}$ as input, generating textual answers $\hat{Y}_{ans}$, which include textual explanations $\hat{Y}_{text}$ and multi-object \texttt{<SEG>} tokens $\hat{Y}_{\texttt{<SEG>}}$. Specifically, each output textual answer consists of a paragraph of textual explanations accompanied by a set of \texttt{<SEG>} tokens, where the number of \texttt{<SEG>} tokens equals the number of 3D objects indicated in the input text query. In each answer, the 3D object's name is followed by a \texttt{<SEG>} token. For example, the LLM might generate the following answer: ``\texttt{You can use the spacious sofa <SEG> for seating, positioned near a central table <SEG> for drinks and snacks, while additional chairs <SEG> provide extra seating options for guests.}" Each \texttt{<SEG>} token indicates a request for a point cloud segmentation mask.

\paragraph{Object-Specific Embedding Extraction.} 
After obtaining the multi-object \texttt{<SEG>} tokens $\hat{Y}_{\texttt{<SEG>}}$ in the output textual answer $\hat{Y}_{ans}$, we extract the point cloud embeddings generated by the multimodal LLM for each \texttt{<SEG>} token and feed them into the 3D Decoder to generate the corresponding segmentation masks. As illustrated in Figure~\ref{fig:object_specific_feature_extraction}, each point cloud $P$ is associated with $N$ textual answers $\hat{Y}_{ans}$, and we illustrate the process with one textual answer $\hat{y}^{n}_{ans}$ for clarity. The LLM's textual answer includes multiple \texttt{<SEG>} tokens, and we use a multi-seg index list $I_{seg}$ to record the positions of these tokens in the tokenized output. 
$I_{seg}$ is obtained from the ground-truth answer ${y}^{n}_{ans}$ during training, and the predicted answer $\hat{y}^{n}_{ans}$ during inference. The multi-seg index list $I_{seg}$ is defined by:
\begin{equation}
    I_{seg} = \{id_0, id_1, ..., id_S \},
\end{equation}
where $S$ denotes the number of \texttt{<SEG>} tokens in the textual answer, and $id$ denotes indices of \texttt{<SEG>} tokens. $I_{seg}$ is then used to retrieve the corresponding point cloud embeddings $F_{seg}$ from the LLM’s hidden states. The process can be formulated as follows:
\begin{equation}
    F_{seg} = \{f_i, i \in I_{seg} \},
\end{equation}
where $f_i$ is the $i$-th point cloud embedding in the LLM’s hidden states.

For the example in Figure~\ref{fig:object_specific_feature_extraction}, the 3rd and 5th hidden embeddings corresponding to the table and chair are extracted, matching the first and second \texttt{<SEG>} tokens in the answer. Each \texttt{<SEG>} token corresponds to a 3D object in $\hat{y}^{n}_{ans}$, allowing us to obtain the point cloud embeddings from the LLM for subsequent segmentation.

\renewcommand\arraystretch{1.0}

\begin{table*}[ht]

\centering
\setlength{\tabcolsep}{34pt}
\begin{tabular}{l|c|cc}
\hline
Method       & Venue            & cIoU    & gIoU                 \\ \hline\hline
\rowcolor{gray!20} \multicolumn{4}{l}{3D Segmentation Methods} \\ \hline
OpenScene~\cite{peng2023openscene}    &  CVPR 23  & 7.69 & 9.52  \\
PLA~\cite{ding2023pla}   &  CVPR 23    & 10.76 & 10.27 \\
RegionPLC~\cite{yang2024regionplc}  & CVPR 24   & 10.82 & 11.06  \\
3D-STMN~\cite{wu20243d}     &   AAAI 24   &  19.37  &   20.70                 \\ 
RefMask3D~\cite{he2024refmask3d}    & MM 24        & 20.53  &    22.81                \\ 
MDIN~\cite{wu20243dgres} & MM 24         & 23.04  &        23.49            \\
\hline
\rowcolor{gray!20} \multicolumn{4}{l}{3D General Scene Understanding Methods} \\ \hline
3D-LLM~\cite{hong20233d}     & NeurIPS 23       & 21.36  & 22.49                       \\  
3D-VisTA~\cite{zhu20233d}   & ICCV 23  & 23.95  & 24.77                   \\   \hline 
\rowcolor{gray!20} \multicolumn{4}{l}{3D Reasoning Segmentation Methods} \\ \hline
Reason3D~\cite{huang2025reason3d}     &   3DV 25    &  \underline{25.12} &       \underline{25.90}               \\ 
\textbf{MORE3D (Ours)}   &    -   & 
\textbf{30.19} & \textbf{32.01}  \\ \hline
\end{tabular}
\caption{
    Benchmarking on the ReasonSeg3D validation set. Best in bold, second underlined.
}
\label{tab:sota_comparison}
\end{table*}

\subsection{Explainability}

The LLM-generated textual answers $\hat{Y}_{ans}$
contain explanations of the implicit intention of the user's questions. 
In addition, the answer includes the descriptions of 3D spatial relations of multiple objects which provide useful guidance for identifying these objects.
Unlike most existing reasoning-based segmentation methods~\cite{wang2024llm, yan2024visa, he2024segpoint, huang2025reason3d} that just output statements like \texttt{"It is <SEG>"} without further explanation, our generated answer integrates detailed explanations about 3D spatial relations to enhance segmentation. The ground truth answers $Y_{ans}$ also incorporate these 3D spatial relations and are used to supervise the generated output textual answers, guiding the model to capture spatial information effectively. The textual answer loss is defined by:
\begin{equation}
    L_{ans} = CE(Y_{ans}, \hat{Y}_{ans}),
    \label{equ:l_ans}
\end{equation}
where $CE$ denotes the cross-entropy loss.

\subsection{Training Objectives}

For point cloud segmentation, we optimize each predicted point cloud segmentation mask as follows:
\begin{equation}
    L_{mask} = BCE(M, \hat{M}) + DICE(M, \hat{M}),
    \label{equ:l_mask}
\end{equation}
where $M$ denotes the ground truth segmentation mask, $\hat{M}$ denotes the predicted segmentation mask, $BCE$ denotes the binary cross-entropy loss and $DICE$ denotes the Dice loss~\cite{milletari2016v}. We use the cross-entropy loss to supervise the point cloud classification as follows:
\begin{equation}
    L_{cls} = CE(C, \hat{C}),
\end{equation}
where $C$ is the ground truth classification label and $\hat{C}$ is the classification prediction.

The proposed model is trained end-to-end with a textual answer loss $L_{ans}$ for textual answer generation, a mask loss $L_{mask}$ for point cloud segmentation mask prediction, and a classification loss $L_{cls}$ for point cloud classification. All three loss terms are assigned equal weights, and the overall loss function is formulated as follows:
\begin{equation}
    L = L_{ans} + L_{mask} + L_{cls}.
\end{equation}

\section{Experiment}
\subsection{Experimental Settings}

\paragraph{Evaluation Metrics.}
Following prior studies on reasoning segmentation~\cite{lai2024lisa, ren2024pixellm, wang2024llm}, we adopt two evaluation metrics including cumulative IoU (cIoU) and generalized IoU (gIoU). cIoU is computed as the cumulative intersection over cumulative union, while gIoU is the average Intersection-over-Union (IoU) across all samples.

\paragraph{Implementation Details.}
We conduct experiments on one NVIDIA V100 GPU (32 GB memory) and train the network for 100 epochs with a batch size of 1, with a total training time of around 1 day. We use the Adam~\cite{kingma2015adam} optimizer with the initial learning rate of $1 \times 10^{-4}$. We adopt LLaMA-7B~\cite{touvron2023llama} as our multimodal LLM backbone and LoRA~\cite{hu2021lora} to perform efficient fine-tuning. All experiments are conducted on the proposed ReasonSeg3D dataset.

\begin{figure*}[ht]
    \centering
    \includegraphics[width=1.0\linewidth]{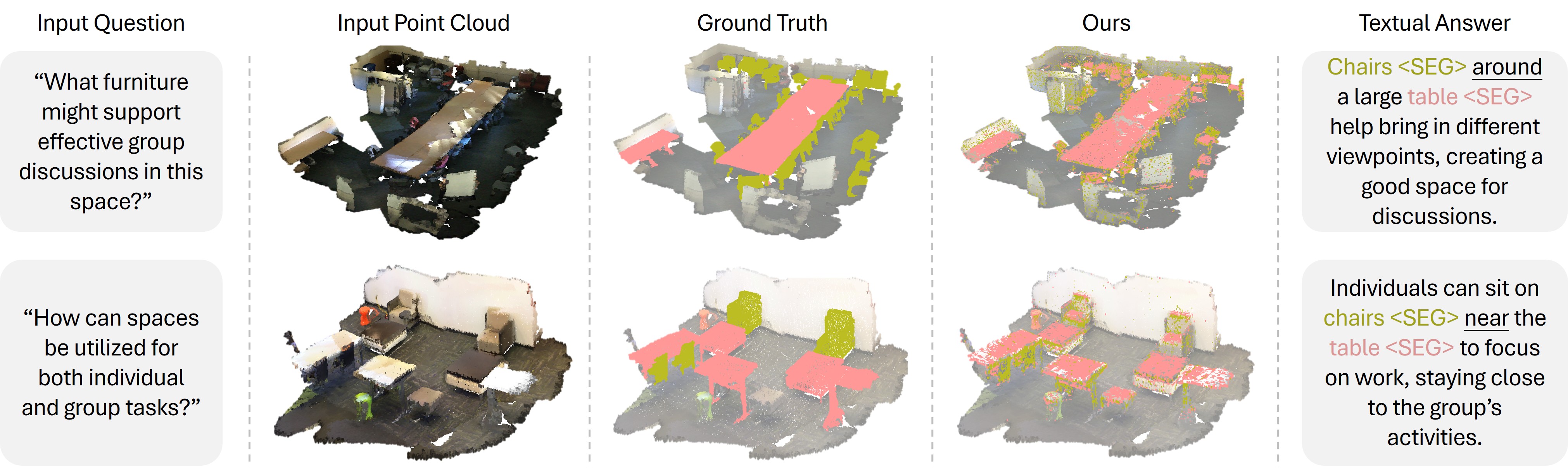}

    \caption{
    Segmentation visualization over the ReasonSeg3D validation set. Each case presents a user input question, the corresponding input point cloud, the ground-truth segmentation, and the prediction by the proposed MORE3D. Green indicates chairs, and pink indicates tables. Underlined text denotes 3D spatial relations. Best viewed in color and zoom-in.
    }
    \label{fig:visualization}
\end{figure*}

\begin{table}[]

\centering
\setlength{\tabcolsep}{15pt}
\begin{tabular}{c|c|cc}
\hline
Index        &  Manner      & cIoU    & gIoU               \\ \hline\hline
1         & Unified & 29.06 & 30.43     \\
2          & Separated & \textbf{30.19} & \textbf{32.01}  \\ \hline
\end{tabular}
\caption{
    Ablation study on different approaches of prediction in the 3D Decoder.
}
\label{tab:prediction_manner}
\end{table}

\begin{table}[]

\centering
\setlength{\tabcolsep}{11pt}
\begin{tabular}{c|cc|cc}
\hline
Index        &  $L_{ans}$ & $L_{mask}$    & cIoU    & gIoU                    \\ \hline\hline
1         &  &   & 13.72 & 14.73 \\
2         & \checkmark &   & 15.23 & 16.14  \\
3        &   & \checkmark & 23.30 & 28.91  \\
4          & \checkmark  & \checkmark & \textbf{30.19} & \textbf{32.01}\\ \hline
\end{tabular}
\caption{
    Ablation study of the loss functions on ReasonSeg3D validation set.
}
\label{tab:loss_ablation}
\end{table}

\subsection{Benchmarking with Existing Methods}

We benchmark MORE3D with state-of-the-art methods, including 3D segmentation~\cite{peng2023openscene, ding2023pla, yang2024regionplc, wu20243d, he2024refmask3d, wu20243dgres}, general scene understanding~\cite{hong20233d, zhu20233d}, and reasoning segmentation~\cite{huang2025reason3d} approaches, on the ReasonSeg3D validation set. All compared methods are trained and evaluated on ReasonSeg3D to ensure a fair comparison. As Table~\ref{tab:sota_comparison} shows, our method achieves superior performance across all evaluation metrics. Specifically, OpenScene~\cite{peng2023openscene}, PLA~\cite{ding2023pla}, and RegionPLC~\cite{yang2024regionplc} are designed to comprehend input words or phrases based on CLIP-based language models. They lack the ability to reason user intention from input questions and to correctly generate answers targeting multiple corresponding objects. Consequently, the predicted 3D segmentation masks by these methods cannot be aligned correctly with ground-truth 3D segmentation masks, resulting in low cIoU and gIoU. Moreover, 3D-STMN~\cite{wu20243d} and RefMask3D~\cite{he2024refmask3d} are tailored for single-object segmentation and therefore cannot handle multiple objects, while MDIN~\cite{wu20243dgres} exhibits weaker reasoning ability for interpreting implicit human intention. Additionally, 3D-LLM~\cite{hong20233d} and 3D-VisTA~\cite{zhu20233d}, designed for general 3D tasks, are re-implemented on our proposed 3D reasoning segmentation task for comparison. Furthermore, Reason3D~\cite{huang2025reason3d}, as a 3D reasoning segmentation approach, is limited to single-object settings and struggles with multi-object reasoning segmentation. In contrast, MORE3D can accurately reason the implicit intention of user questions, predicting 3D segmentation masks that are more consistent with the ground truth.

\paragraph{Qualitative Results.} 
Figure~\ref{fig:visualization} presents qualitative segmentation by the proposed MORE3D with two examples from the ReasonSeg3D validation set. Each example demonstrates the user question, input point cloud, ground-truth segmentation, and the 3D segmentation masks that are predicted by MORE3D. We can observe that MORE3D could accurately predict the 3D segmentation masks that are highly aligned with the ground-truth segmentation masks. 
Specifically, in the top example, despite many chairs closely surrounding the central long table, MORE3D can correctly distinguish and segment them. In the bottom example, despite the similar appearance and size of tables and chairs, MORE3D can distinguish and segment them precisely. The visualization indicates that MORE3D can comprehend the implicit intention behind the user questions and segment multiple objects accurately.

\subsection{Ablation Study}

We conduct ablation studies on the ReasonSeg3D validation set to evaluate our designs. Specifically, we examine MORE3D with the approach for the way of prediction, and the loss design.

\paragraph{Prediction Approaches.} 
We examine the effectiveness of different prediction approaches including unified and separated approaches for the prediction of 3D segmentation masks and classification. Table~\ref{tab:prediction_manner} shows experimental results. For the unified approach, the 3D segmentation masks and classification results are predicted by a single network instead of two separate network branches. This results in lower performance due to the increased learning burden when a single network must learn to perform segmentation and classification simultaneously. In contrast, the separate approach as illustrated in Figure~\ref{fig:overall_architecture} assigns each task to an independent network branch to learn. This allows each branch to focus on its respective task, which lowers the learning burden and improves performance clearly.

\paragraph{Loss Functions.} 
We examine the impact of the textual answer loss $L_{ans}$ and the mask loss $L_{mask}$ in Equations~\ref{equ:l_ans} and~\ref{equ:l_mask}, where $L_{ans}$ supervises the textual answers output by the multimodal LLM and $L_{mask}$ supervises the 3D point cloud segmentation prediction. As Table~\ref{tab:loss_ablation} shows, without $L_{ans}$ and $L_{mask}$, the performance drops greatly due to the lack of supervision from both ground-truth textual answers and segmentation masks, which hinders the model from generating appropriate answers and accurate 3D segmentation masks. When either $L_{ans}$ or $L_{mask}$ is used, the performance improves consistently due to partial supervision from the ground-truth textual answers or the segmentation masks. When both $L_{ans}$ and $L_{mask}$ are employed, the reasoning ability for generating answers and the capability of predicting 3D segmentation masks of MORE3D are both improved by large margins, demonstrating the effectiveness and synergies of the two designed losses.

\section{Conclusion}

This paper presents a novel multi-object 3D reasoning segmentation task, producing both 3D segmentation masks and textual explanations with rich 3D spatial relations among multiple objects within complex 3D scenes. To this end, we develop ReasonSeg3D, a large-scale benchmark featuring rich 3D spatial relations integrated into question-answer pairs, designed to evaluate multi-object 3D reasoning segmentation effectively. On top of this, we propose MORE3D, a technique for multi-object 3D reasoning segmentation with textual explanations, demonstrating strong reasoning abilities in response to user questions. Extensive experiments validate the effectiveness of MORE3D. 
Future work will focus on exploring our work in more challenging 3D environments to improve applicability in real-world scenarios.

\bibliography{aaai2025}

\end{document}